\pdfoutput=1
\documentclass[11pt]{article}

\usepackage[]{EACL2023}

\usepackage{times}
\usepackage{latexsym}
\usepackage{graphicx}
\usepackage[acronym]{glossaries}

\usepackage[T1]{fontenc}
\usepackage[utf8]{inputenc}

\usepackage{microtype}

\usepackage{inconsolata}

\title{\acrshort{t5} for Hate Speech, Augmented Data and Ensemble}

\author{\\Tosin Adewumi, Sana Sabah Sabry, Nosheen Abid, Foteini Liwicki \& Marcus Liwicki\\
ML Group, EISLAB, Luleå University of Technology, Sweden \\
  \texttt{firstname.lastname@ltu.se} }

\newacronym{nlp}{NLP}{Natural Language Processing}
\newacronym{ner}{NER}{Named Entity Recognition}
\newacronym{sa}{SA}{Sentiment Analysis}
\newacronym{bow}{BoW}{bag-of-words}
\newacronym{cbow}{CBoW}{continuous Bag-of-Words}
\newacronym{sltc}{SLTC}{Swedish Language Technology Conference}
\newacronym{ann}{ANN}{artificial neural network}
\newacronym{nn}{NN}{neural network}
\newacronym{lstm}{LSTM}{Long Short Term Memory Network}
\newacronym{sota}{SoTA}{state-of-the-art}
\newacronym{nlg}{NLG}{Natural Language Generation}
\newacronym{mwe}{MWE}{Multi-Word Expression}
\newacronym{cnn}{CNN}{Convolutional Neural Network}
\newacronym{sw}{SW}{Simple Wiki}
\newacronym{mt}{MT}{Machine Translation}
\newacronym{gdc}{GDC}{Gothenburg Dialog Corpus}
\newacronym{t5}{T5}{Text-to-Text-Transfer Transformer}
\newacronym{roberta}{RoBERTa}{Robustly optimized BERT approach}
\newacronym{bert}{BERT}{Bidirectional Encoder Representations from Transformers}
\newacronym{mcc}{MCC}{Matthews Correlation Coefficient}
\newacronym{ai}{AI}{artificial intelligence}
\newacronym{xai}{XAI}{explainable artificial intelligence}
\newacronym{lime}{LIME}{Local Interpretable Model-agnostic Explanations}
\newacronym{bilstm}{Bi-LSTM}{Bi- Directional Long Short Term Memory Network}
\newacronym{rnn}{RNN}{Recurrent Neural Network}
\newacronym{ml}{ML}{machine learning}
\newacronym{hasoc}{HASOC}{hate speech and offensive content}
\newacronym{olid}{OLID}{offensive language identification dataset}
\newacronym{ig}{IG}{Integrated Gradient}
\newacronym{shap}{SHAP}{SHapley Additive exPlanations}
\newacronym{hs}{HS}{hate speech}
\newacronym{trac}{TRAC}{Trolling, Aggression and Cyberbullying}
\newacronym{hso}{HSO}{hate speech and offensive}
\newacronym{ooc}{OOC}{out-of-class}
\newacronym{os}{OS}{operating system}
\newacronym{lr}{LR}{learning rate}
\newacronym{multiwoz}{MultiWOZ}{Multi-Domain Wizard-of-Oz}
\newacronym{dialogpt}{DialoGPT}{Dialogue Generative Pre-trained Transformer}

\begin{document}
\maketitle
\begin{abstract}
We conduct relatively extensive investigations of  automatic hate speech (\acrshort{hs}) detection using different \acrfull{sota} baselines over 11 subtasks of 6 different datasets.
Our motivation is to determine which of the recent \acrshort{sota} models is best for automatic hate speech detection and what advantage methods like data augmentation and ensemble may have on the best model, if any.
We carry out 6 cross-task investigations.
We achieve new \acrshort{sota} on two subtasks - macro F1 scores of 91.73\% and 53.21\% for subtasks A and B of the \acrshort{hasoc} 2020 dataset, where previous \acrshort{sota} are 51.52\% and 26.52\%, respectively.
We achieve near-\acrshort{sota} on two others - macro F1 scores of 81.66\% for subtask A of the \acrshort{olid} 2019 dataset and 82.54\% for subtask A of the \acrshort{hasoc} 2021 dataset, where \acrshort{sota} are 82.9\% and 83.05\%, respectively.
We perform error analysis and use two \acrfull{xai} algorithms (\acrshort{ig} and \acrshort{shap}) to reveal how two of the models (\acrshort{bilstm} and \acrshort{t5}) make the predictions they do by using examples.
Other contributions of this work are
1) the introduction of a simple, novel mechanism for correcting \acrfull{ooc} predictions in \acrshort{t5}, 2) a detailed description of the data augmentation methods, 3) the revelation of the
poor data annotations in the \acrshort{hasoc} 2021 dataset by using several examples and \acrshort{xai}
(buttressing the need for better quality control), and 4) the public release of our model checkpoints and codes to foster transparency.\footnote{available after anonymity period}

\end{abstract}

\section{Introduction}

Any disparaging remark targeted at an individual or group of persons is usually considered as \acrfull{hs} \cite{nockleby2000hate,brown2017hate}.
It is considered unethical in many countries and illegal in some \cite{brown2017hate,quintel2020self,fortuna2018survey}.
Manual detection of \acrshort{hs} content is a tedious task that can result in delays in stopping harmful behaviour.\footnote{bbc.com/news/world-europe-35105003}
Automatic hate speech detection is, therefore, crucial and has been gaining increasing importance because of the rising influence of social media in many societies.
It will facilitate the elimination/prevention of undesirable characteristics in data, and by extension \acrshort{ai} technologies, such as conversational systems \cite{zhang2020dialogpt,adewumi2021sm}.
\acrshort{hs} examples that may incite others to violence in the \acrfull{olid} data \cite{zampierietal2019} are given in Table \ref{table:olid}.

\begin{table}[h]
\small
\centering
\resizebox{\columnwidth}{!}{%
\begin{tabular}{p{0.07\linewidth} p{0.85\linewidth}}
\hline
id & tweet
\\
\hline
23352 & @USER Antifa simply wants us to k*ll them. By the way. Most of us carry a back up. And a knife
\\
\hline
61110 & @USER @USER Her life is crappy because she is crappy. And she’s threatening to k*ll everyone. Another nut job...Listen up FBI!
\\
\hline
68130 & @USER @USER @USER @USER @USER Yes usually in THOSE countries people k*ll gays cuz religion advise them to do it and try to point this out and antifa will beat you.  No matter how u try in america to help gay in those countries it will have no effect cuz those ppl hate america.
\\
\hline
\end{tabular}
}
\caption{\label{table:olid}
Inciteful examples from the \acrshort{olid} 2019 training set. \footnotesize
(parts of offensive words masked with "*")
}
\end{table}

Short details of the datasets in this work are provided in appendix \ref{datasetsused}.
The datasets were selected based on the important subtasks covered with regards to \acrshort{hs} or abusive language.
The architectures employed include the \acrfull{bilstm}, the \acrfull{cnn}, \acrfull{roberta}-Base, \acrfull{t5}-Base, where the last two are pretrained models from the HuggingFace hub.
As the best-performing baseline model, \acrshort{t5}-Base is then used on the augmentated data for the \acrshort{hasoc} 2021 subtasks A \& B and for an ensemble.
In addition, we compare result from HateBERT, a re-trained \acrshort{bert} model for abusive language detection \cite{caselli-etal-2021-hatebert}.

The rest of this paper is structured as follows: Section~\ref{methodology} explains the methods used in this study.
The results, critical analysis with \acrshort{xai} and discussion are in Section~\ref{results}.
Section~\ref{related_work} provides an overview of \acrshort{hs} and prior work in the field.
Section~\ref{conclude} gives the conclusion and possible future work.

\section{Methodology}
\label{methodology}
All the experiments were conducted on a shared DGX-1 machine with 8 × 32GB Nvidia V100 GPUs.
The \acrfull{os} of the server is Ubuntu 18 and it has 80 CPU cores.
Each experiment is conducted 3 times and the average results computed.
Six is the total number of epochs for each experiment and the model checkpoint with the lowest validation loss is saved and used for evaluation of the test set, where available.
Linear schedule with warm up is used for the \acrfull{lr} adjustment for \acrshort{t5} and \acrshort{roberta}.
Only limited hyperparameters are explored, through manual tuning, for all the models due to time and resource constraints.

Short details about all the models used are discussed in appendix~\ref{models}.
Appendix~\ref{methodsapp} gives more information on the data preprocessing, metrics for evaluation, the ensemble, and cross-task training.
Average time per epoch for training and evaluation on the validation set is 83.52, 7.82 \& 22.29 seconds for the \acrshort{olid}, \acrshort{hasoc} 2020 \&  \acrshort{hasoc} 2021 datasets, respectively.\footnote{
Restrictions (\textit{cpulimit}) were implemented to avoid server overloading, in fairness to other users.
Hence, average time for the test sets ranges from 2 to over 24 hours.}

\subsection{Solving \acrshort{ooc} Predictions in \acrshort{t5}}

\citet{JMLR:v21:20-074} introduced \acrshort{t5} and noted the possibility of \acrshort{ooc} predictions in the model.
This is when the model predicts text (or empty string) seen during training but is not among the class labels.
This issue appears more common in the initial epochs of training and may not even occur
sometimes.
We experienced this challenge in the two libraries we attempted to develop with.\footnote{HuggingFace \& Simple Transfromers}
In order to solve this, first we introduced integers (explicitly type-cast as string) for class labels, which appear to make the model predictions more stable.
The issue reduced by about 50\% in pilot studies, when they occur.
For example, for the \acrshort{hasoc} datasets, we substituted "1" and "0" for the labels "NOT" and "HOF", respectively.
As a second step, a simple correction we introduced is to replace the \acrshort{ooc} prediction (if it occurs) with the label of the largest class in the training set.

\subsection{Data Augmentation}
\label{dataaug}
The objective of data augmentation is to increase the number of training data samples in order to improve performance of models on the evaluation set \cite{feng-etal-2021-survey}.
We experimented with 2 techniques: 1) word-level deletion of the start and end words per sample and 2) conversational \acrshort{ai} text generation (Table \ref{table:hasaug}).
Our work may be the first to use conversational \acrshort{ai} for data augmentation.
It doubles the amount of samples and provides diversity.
The average new words generated per sample prompt is around 16 words.
More details about the 2 techniques are found in appendix \ref{augment}.

\begin{table}[h]
\small
\centering
\resizebox{\columnwidth}{!}{%
\begin{tabular}{p{0.13\linewidth} p{0.8\linewidth}}
\hline
Type & Sample
\\
\hline
original & Son of a *** wrong "you're"
\\
augmented & son of a *** wrong youre No, that's Saint Johns Chop House. I need a taxi to take me from the hotel to the restaurant, leaving the first at 5:45.
\\ % \hline
\hline
original & SO EXCITED TO GET MY CovidVaccine I hate you covid!
\\
augmented & so excited to get my covidvaccine i hate you covid You should probably get that checked out by a gastroenterology department.
\\
\hline
original & ModiKaVaccineJumla Who is responsible for oxygen? ModiResign Do you agree with me? â¤ï¸ Don't you agree with me?
\\
augmented & modikavaccinejumla who is responsible for oxygen modiresign do you agree with me âï dont you agree with me Yes, I definitely do not want to work with them again. I appreciate your help..
\\
\hline

\end{tabular}
}
\caption{\label{table:hasaug}
Original and conversational \acrshort{ai}-augmented examples from the \acrshort{hasoc} 2021 dataset. \footnotesize (offensive words masked with "***")
}
\end{table}

\begin{table*}[h!]
\scriptsize %footnotesize
\centering
\begin{tabular}{lcccc}
\hline
\textbf{Task} &
\multicolumn{2}{c}{\textbf{Weighted F1 (\%)}} &
\multicolumn{2}{c}{\textbf{Macro F1(\%)}}
\\
%\hline
 & Dev (sd) & Test (sd) & Dev (sd) & Test (sd)
\\
\hline
\textbf{Bi-LSTM} & \multicolumn{4}{c}{}
\\
\hline
OLID A &  79.59 (0.89) & 83.89 (0.57) &  78.48 (1.52) & 79.49 (0)
\\
OLID B &  82.50 (1.70) & 83.46 (0) &  46.76 (0) & 47.32 (0)
\\
OLID C &  49.75 (3.95) & 43.82 (9.63) &  35.65 (2.81) & 36.82 (0)
\\
Hasoc 2021 A &  78.05 (0.85) &  78.43 (0.84) &  77.99 (1.79) &  77.19 (0)
\\
Hasoc 2021 B &  50.65 (1.34) &  52.19 (1.95) & 43.19 (2.09) & 42.25 (0)
\\
\hline
\textbf{CNN} & \multicolumn{4}{c}{}
\\
\hline
OLID A &  79.10 (0.26) & 82.47 (0.56)  &  77.61 (0.39) & 78.46 (0)
\\
OLID B &  82.43 (0.49) & 83.46 (0) & 46.76 (0) & 47.88 (0)
\\
OLID C &  47.54 (1.36) & 38.09 (3.91) &  35.65 (0) & 36.85 (0)
\\
Hasoc 2021 A &  77.22 (0.52) &  77.63 (0.70) &   74.28 (0.58) &   75.67 (0)
\\
Hasoc 2021 B &  55.60 (0.61) &  59.84 (0.41) & 50.41 (0.41) & 54.99 (0)
\\
\hline
\textbf{RoBERTa} & \multicolumn{4}{c}{}
\\
\hline
OLID A & 82.70 (0.55) & 84.62 (0) &  80.51 (0.76) & 80.34 (0)
\\
OLID B & 82.70 (0.13) & 83.46 (0) & 46.76 (0.04) & 47.02 (0)
\\
Hasoc 2021 A & 79.9 (0.57) & 76.2 (0) & 77.77 (0.75) & 74 (0)
\\
\hline
\textbf{T5-Base} & \multicolumn{4}{c}{}
\\
\hline
OLID A & 92.90 (1.37) & 85.57 (0) & 92.93 (1.42) & 81.66 (0)
\\
OLID B & 99.75 (0.43) & 86.81 (0) & 99.77 (0.44) & 53.78 (0)
\\
OLID C & 58.35 (1.22) & 54.99 (0) & 33.09 (0.76) & 43.12 (0)
\\
Hasoc 2021 A & 94.60 (1.98) & 82.3 (0) & 94.73 (5.26) & 80.81 (0)
\\
Hasoc 2021 B & 65.40 (0.82) & 62.74 (0) & 62.43 (6.32) & 59.21 (0)
\\
\hline
\textbf{\cite{zampieri2019semeval}} best scores& \multicolumn{4}{c}{}
\\
\hline
OLID A &  & &  & \textbf{82.90} (-)
\\
OLID B &  &  &  & \textbf{75.50} (-)
\\
OLID C &  &  & & \textbf{66} (-)
\\
\hline
\end{tabular}
%}
\caption{\label{tab:compare} Mean scores of model baselines for different subtasks. \footnotesize{(sd: standard deviation; bold values are best scores for a given task; '-' implies no informaton available)}}
\end{table*}

\begin{table*}[h!]
\scriptsize %footnotesize
\centering
\begin{tabular}{lcccc}
\hline
\textbf{Task} &
\multicolumn{2}{c}{\textbf{Weighted F1 (\%)}} &
\multicolumn{2}{c}{\textbf{Macro F1(\%)}}
\\
%\hline
 & Dev (sd) & Test (sd) & Dev (sd) & Test (sd)
\\
\hline
\textbf{T5-Base} & \multicolumn{4}{c}{}
\\
\hline
Hasoc 2020 A & 96.77 (0.54) & 91.12 (0.2) & 96.76 (0.54) & \textbf{91.12} (0.2)
\\ % \hline
Hasoc 2020 B & 83.36 (1.59) & 79.08 (1.15) & 56.38 (5.09) & \textbf{53.21} (2.87)
\\
\hline
\textbf{T5-Base+Augmented Data} & \multicolumn{4}{c}{}
\\
\hline
Hasoc 2021 A & 95.5 (3.27) & 83 (0) & 92.97 (2.20) & 82.54 (0)
\\
Hasoc 2021 B & 64.74 (3.84) & 66.85 (0) & 65.56 (1.48) & 62.71 (0)
\\
\hline
\textbf{Ensemble} & \multicolumn{4}{c}{}
\\
\hline
Hasoc 2021 A &  & 80.78 (0) &  & 79.05 (0)
\\
%Hasoc 2021 B & () &  () &  () & () \\ 
\hline
\textbf{\cite{mandl2020overview}} best scores & \multicolumn{4}{c}{\textbf{}}
\\
\hline
Hasoc 2020 A &  &  &  &  51.52 (-)
\\
Hasoc 2020 B &  &  &  & 26.52 (-)
\\
\hline
\textbf{\cite{mandl2021overview}} best scores & \multicolumn{4}{c}{\textbf{}}
\\
\hline
Hasoc 2021 A &  &  &  &  \textbf{83.05} (-)
\\
Hasoc 2021 B &  &  &  & \textbf{66.57} (-)
\\
\hline

\end{tabular}
%}
\caption{\label{tab:comparet5} \acrshort{t5} variants' mean scores over \acrshort{hasoc} data. \footnotesize{(sd: standard deviation; bold values are best scores for a given task; '-' implies no informaton available)}}
\end{table*}

\section{Results and Discussion}
\label{results}
Tables \ref{tab:compare}, \ref{tab:comparet5} and \ref{table:crosst} (Appendix \ref{sec:appendix}) show baseline results, additional results using the best model: \acrshort{t5}, and the cross-task with \acrshort{t5}, respectively.
Table \ref{tab:compare2} (Appendix \ref{sec:appendix}), shows results for other datasets and the HateBERT model \cite{caselli-etal-2021-hatebert}.
The HatEval task is the only comparable one in our work and that by \citet{caselli-etal-2021-hatebert}.

\paragraph{The Baselines}:
The Transformer-based models (\acrshort{t5} and \acrshort{roberta}) generally perform better than the other baselines (\acrshort{lstm} and \acrshort{cnn}) \cite{zampieri2019semeval}, except for \acrshort{roberta} on the \acrshort{olid} subtask B and \acrshort{hasoc} 2021 subtask A.
The \acrshort{t5} outperforms \acrshort{roberta} on all tasks.
Based on the test set results, the \acrshort{lstm} obtains better results than the \acrshort{cnn} in the \acrshort{olid} subtasks A, \acrshort{hasoc} 2020 subtask A, and \acrshort{hasoc} 2021 subtask A while the \acrshort{cnn} does better than it on the others.
\paragraph{\acrshort{t5} Variants \& Augmentation}:
The \acrshort{t5}-Base model achieves new best scores on the \acrshort{hasoc} 2020 subtasks.
The augmented data, using the conversational \acrshort{ai} technique, improves results on \acrshort{hasoc} 2021.\footnote{The first technique is not reported because there was no improvement.
This may be because the number of total samples is smaller than that of the conversational \acrshort{ai} technique.}

\subsection{The Ensemble}
The ensemble macro F1 result (79.05\%) is closer to the \acrshort{t5}-Base result (80.81\%) and farther from the \acrshort{roberta} result (74\%).
The deciding factor is the \acrshort{t5}-Small.
Hence, a voting ensemble may not perform better than the strongest model in the collection if the other models are weaker at prediction.

\subsection{Cross-Task Training}
We obtain new \acrshort{sota} result (91.73\%) for the \acrshort{hasoc} 2020 subtask A after initial training on the \acrshort{olid} subtask A.
The reason we outperform the previous \acrshort{sota} is that they used an \acrshort{lstm} with GloVe embeddings \cite{mandl2020overview}, instead of a pretrained deep model with the attention mechanism \cite{bahdanau2015neural} that gives transfer learning advantage.
The p-value (p < 0.0001) obtained for the difference of two means of the two-sample t-test is smaller than the alpha (0.05), showing that the results are statistically significant.

\subsection{\acrshort{hasoc} 2021 Annotation Issues}
Inspection of some of the samples predicted by the \acrshort{t5} model reveal challenges with the quality of data annotation in the \acrshort{hasoc} 2021 dataset.
Table \ref{table:hasissues} (Appendix \ref{sec:appendix}) gives several (10) examples of tweets incorrectly labeled as "NOT" ('1') by the annotators but which are clearly offensive (HOF ('0')), in our view, and are also correctly predicted as such by the model.
More cases like these exist within the dataset than shown in the table.
This issue makes a strong case for having better quality control (QC) with data annotation, given the possible implications, including the poor assessments that may result from the competitions organized using such dataset.
We provide \acrshort{shap} explanations of the \acrshort{t5} model predictions for some of these suspicious examples (Appendix \ref{incorrect}).

\subsection{Error Analysis}
\begin{figure}[h]
\centering
\includegraphics[width=0.4\textwidth]{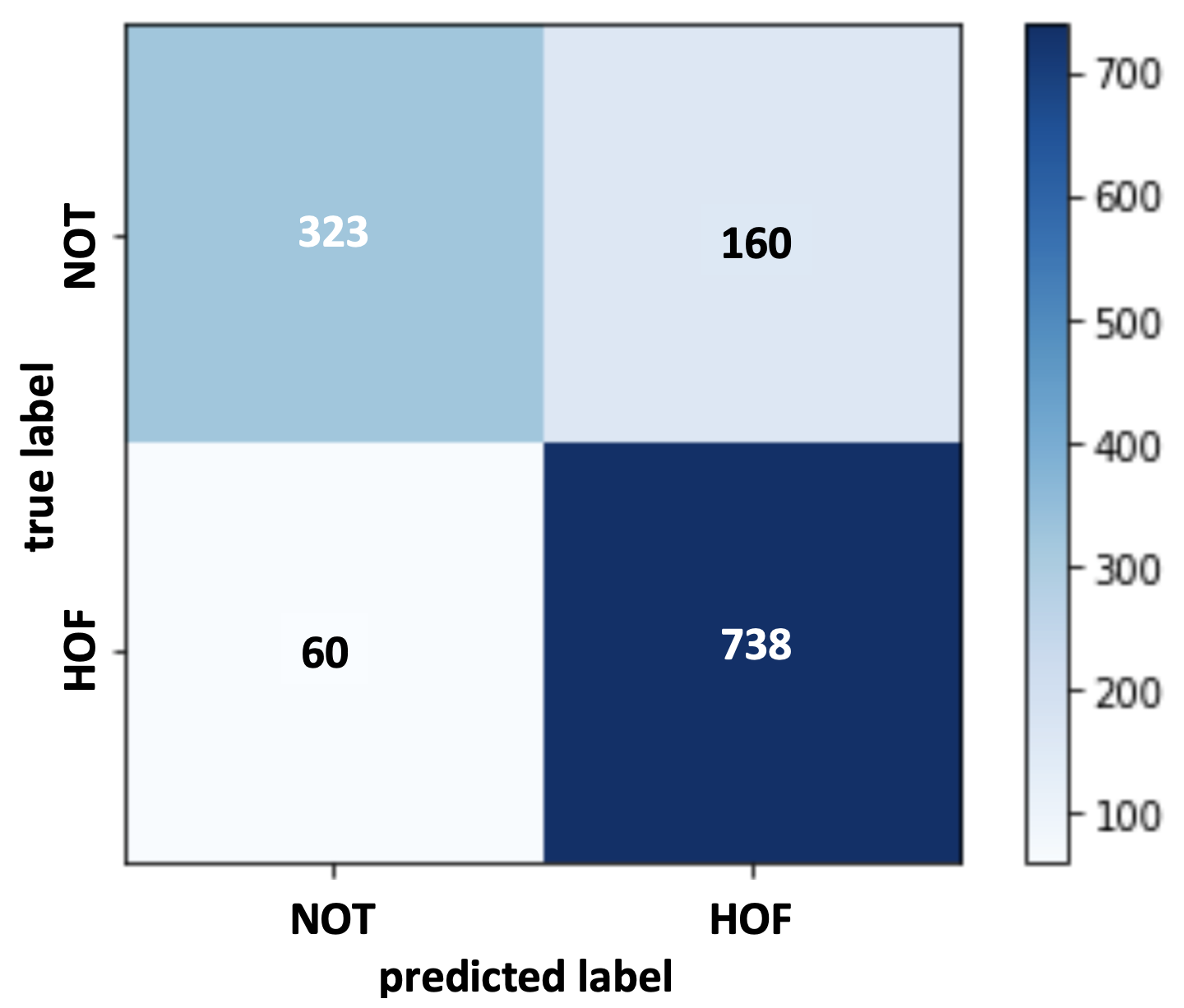}
\caption{Confusion matrix of \acrshort{t5} on Hasoc 2021 test set}
\label{fig:cmat}
\end{figure}

The confusion matrix for the \acrshort{t5} on \acrshort{hasoc} 2021 is given by Figure \ref{fig:cmat}.
It reveals that 33\% (160) of the "NOT" class (not offensive) was misclassified as offensive while only 8\% (60) of the "HOF" (hate or offensive) was misclassified as "NOT".
The higher percentage of misclassification for the "NOT" class is very likely due to the fact that the training set is imbalanced, as there are more "HOF" samples (2,251) compared to "NOT" (1,207).
Hence, the model is better at identifying samples of "HOF".

\subsection{Explainable Artificial Intelligence (\acrshort{xai})}
\acrshort{xai} helps us understand how a model arrives at a prediction and identify any incompleteness in the model \cite{doshi2017towards}.
This can add to the justification for using \acrshort{ml} models and the trust in their predictions.
In this study, rather than compare two \acrshort{xai} algorithms on one model, we focus on separate explanations from two \acrshort{xai} algorithms on two different models, using the same examples from the \acrshort{hasoc} 2021 test set subtask A (Appendix \ref{sec:appendix}, Table \ref{xai_samples}).
The \acrshort{xai} algorithms are \acrshort{ig} and \acrshort{shap}, which are discussed in detail in appendix \ref{sec:xaialg} with examples.

\section{Related Work}
\label{related_work}

Significant efforts have gone into addressing automatic \acrshort{hs} detection \cite{davidson2017automated,matthew2020hatexplain}.
\citet{zampierietal2019} extended the \acrshort{olid} dataset to annotate the distinction between explicit and implicit
messages.
\citet{caselli2020feel} performed cross-domain experiments on HatEval \cite{basile2019semeval}.
\citet{mutanga2020hate} experimented with different Transformer-based architectures using only the \acrshort{hso} dataset.
However, their preprocessing approach, which involves removing low frequency words, may result in newly introduced hate terms escaping detection.

The Transformer architecture by \citet{vaswani2017attention} has been very influential in recent progress with various \acrshort{nlp} tasks.
The attention mechanism on which it is based makes it possible for it to handle long-term dependencies \cite{bahdanau2015neural,vaswani2017attention}.
Hence, Transformer-based models have gained increased attention in \acrshort{hs} detection and classification \cite{mutanga2020hate,matthew2020hatexplain,kovacs2021challenges,elsafoury2021does}.
Despite the introduction of these models, there seems to be a gap where recent \acrshort{sota} models are not compared across many \acrshort{hs} datasets.
We address that in this work.

\section{Conclusion}
\label{conclude}
In this study, we solve the \acrshort{ooc} problem in \acrshort{t5} using a simple two-step approach, demonstrate the benefits of data augmentation through conversational \acrshort{ai} and cross-task training for automatic \acrshort{hs} detection.
We achieve new \acrshort{sota} results on the \acrshort{hasoc} 2020 subtasks A and B.
We also achieve near-\acrshort{sota} results for both the subtask A of the \acrshort{olid} 2019 and \acrshort{hasoc} 2021 datasets.
We reveal, with examples and \acrshort{xai}, the shortcomings of the \acrshort{hasoc} 2021 dataset and make a case for better quality control with data annotation.
\acrshort{ig} and \acrshort{shap} are also used to explain predictions of some of the same examples from the \acrshort{hasoc} 2021 dataset.
Future work that compares performance with models, which are pretrained on large volumes of tweet, such as BERTweet \cite{nguyen-etal-2020-bertweet}, may be a worthwhile investigation.
Releasing our source codes and model checkpoints provides the opportunity for the community to reproduce our results and foster transparency.

\section*{Limitations}
The datasets used in this study are all in the English language.
The results are, therefore, limited to the English language.
It is unclear how the models will perform with other languages.
Many of the datasets are also based on tweets, which are usually short.
Hence, there might be low scalability of the models to long text.
Furthermore, none of the models has 100\% performance on the short tweets.
Also, all the models were trained on GPU and this requirement is necessary to train the models to speed up training time.

\section*{Ethics Statement}
The results obtained in this work are factual and reproducible.
Conscious effort was made by the authors to avoid harm in the presentation of this study though some data samples contain offensive or \acrlong{hs} content.
For example, many of the offensive words or characters have been masked with "*".
Although we have used \acrshort{xai}, we acknowledge it may lead to the potential risk of undue trust in models and therefore provided more than one \acrshort{xai} algorithm explanations for more than one model.
In fairness to other users of the shared system where the experiments were run, restrictions (\textit{cpulimit}) were implemented during the experiments to avoid overloading the server.

\bibliography{anthology,custom}
\bibliographystyle{acl_natbib}

\appendix
\section*{Appendix}

\section{Data}
\label{datasetsused}

    Following are the datasets considered in this work:

\begin{enumerate}
    \item \acrshort{hasoc} 2020
    
    The English dataset is composed of 3,708 tweets for training and 1,592 for testing.
    The dataset includes the following subtasks: 1) task\_1 (A), which identifies hate and offensive text and 2) task\_2 (B), which is a further classification for the previous task to categorize the hateful and offensive content into either hate content (HATE), offensive (OFFN) or profane (PRFN). \citet{mandl2020overview} collected the dataset and used a trained SVM classifier and human judgment to label the data. 
    
    \item \acrshort{hasoc} 2021
    
    This third edition of \acrshort{hasoc} \citet{mandl2021overview} provided another set of tweets dataset with the same subtasks as \acrshort{hasoc} 2020.
    The English dataset consists of 3,843 training samples and 1,281 samples in the test set.
    The dataset has Covid-related topics since the data was gathered during the Covid-19 pandemic.
    10\% of the training set is split as the dev set in this work for evaluation after each epoch.
    
    \item HatEval 2019
    
    \citet{basile2019semeval} prepared this dataset of tweets to detect hateful content against women and immigrants.
    It contains 13,000 English tweets, distributed as 9,000 for training, 1,000 for development and 3,000 for testing.
    The dataset includes two subtasks: subtask A identifies the presence of hate speech while subtask B is the average of three binary classification tasks.
    The 3 binary subtasks under subtask B include 1) \acrshort{hs}, 2) whether the hate speech targets group of people or an individual (TR), and if the \acrshort{hs} contains aggressive content or not (AG).

    \item \acrshort{olid} 2019
    
    The SemEval 2019 shared task 6 dataset is based on the \acrshort{olid} dataset.
    It has 14,200 annotated English tweets and encompasses the following three sub-tasks: a) offensive language detection, b) categorization of offensive language as to whether it's targeted at someone (or a group) or not, and c) offensive language target identification, where distinction is made among individual, group and other entities, like an organisation \cite{zampierietal2019}.
    Crowd-workers performed its data annotation and the original data-split was into training and test sets only.
    As we did with \acrshort{hasoc} 2021, we split 10\% of the training set as the dev set for evaluation after each epoch.
    
    \item \acrshort{hso}
    
    \citet{davidson2017automated} gathered tweets based on a hate speech lexicon and employed crowd-sourcing effort to annotate them.
    They make a distinction between hate speech and offensive language, choosing a narrower definition of hate speech, as opposed to some general view like \citet{zampierietal2019}.
    Three categories are present in the labeled data: hate speech, only offensive language, and neither.
    Of the 24,802 labeled tweets, resulting in the \acrshort{hso} data, 5\% were labeled as containing hate speech while 1.3\% were by unanimous decision.

    \item \acrshort{trac}
    
    \citet{trac2-report} introduced \acrshort{trac} and the second version, in 2020, contains two subtasks.
    Three categories are present in the first subtask: Overtly Aggressive, Covertly Aggressive and Non-aggressive.
    The English version of this task contains 5,000 samples for training and evaluation, just like the Bangla and Hindi versions.
    The second subtask is a binary classification to identify gendered or non-gendered text.
    Our focus was on the first subtask only in this work.
    \citet{elsafoury2021timeline} distinguish this dataset from other \acrshort{hs} datasets. However, they also acknowledge that there are some similarities (like abusive language) between aggression and \acrshort{hs}.
    It is based on this that we selected the dataset.
\end{enumerate}

\section{Models}
\label{models}

\subsection{\acrshort{bilstm}}
The \acrshort{bilstm} is one form of \acrfull{rnn}\cite{hochreiter1997long}. %\acrshort{rnn} is used with sequential data and can capture long-term dependencies in text. 
It is an improved variant of the vanilla \acrshort{rnn}.
Its input text flows forward and backwards, thereby providing more contextual information, thereby improving the network performance \cite{graves2005framewise}.
We used 2 bi-directional layers and a pretrained Glove \cite{pennington2014glove} word embeddings of 100 dimensions.
We also applied a dropout layer to prevent overfitting.
This model has 1,317,721 parameters.
Word and subword embeddings have been shown to improve performance of downstream tasks \cite{mikolov2013distributed,pennington2014glove,adewumi2020word2vec}.

\subsection{\acrshort{cnn}}
The \acrshort{cnn} is common in computer vision or image processing. \cite{kim-2014-convolutional} shows the effectiveness of \acrshort{cnn} to capture text local patterns on different \acrshort{nlp} tasks.  
Both the \acrshort{bilstm} and \acrshort{cnn} architectures are used as feature-based models, where for each tweet, we computed embeddings using pre-trained Glove, before using the embeddings as an input to the baseline model.
The \acrshort{cnn} model is composed of 3  convolution layers with 100 filters each. The filter size for the first layer is $2\times100$, the filter size for the second layer is $3\times100$ and $4\times100$ for the third layer.
We use ReLU activation function and max-pooling after each convolution layer.
We perform dropout for regularization.
The total trainable  parameters  for the \acrshort{cnn} are 1,386,201

\subsection{\acrshort{roberta}}
\acrshort{roberta} is based on the replication study of \acrshort{bert}.
It differs from \acrshort{bert} in the following ways: (1) training for longer over more data, (2) removing the next sentence prediction objective, and (3) using longer sequences for training \cite{liu2019roberta}.
The base version of the model, which we use,  has 12 layers and 110M parameters.
For our study, we use a batch size of 32, initial learning rate of 1e-5, and maximum sequence length of 256.
We restricted the number of tasks to only binary tasks for this model.

\subsection{\acrshort{t5}}
The \acrshort{t5} \cite{JMLR:v21:20-074} is based on the transformer architecture by \citet{vaswani2017attention}.
However, a different layer normalization is applied, where there's no additive bias applied and the activations are only rescaled.
Causal or autoregressive self-attention is used in the decoder for it to attend to past outputs.
The \acrshort{t5}-Base model has about twice the number of parameters as that of \acrshort{bert}-Base.
Its has 220M parameters and 12 layers each in the encoder and decoder blocks while the small version has 60M parameters \cite{JMLR:v21:20-074}.
The \acrshort{t5} training method uses teacher forcing (i.e. standard maximum likelihood) and a cross-entropy loss.
\acrshort{t5}-Base required more memory and would not fit on a single V100 GPU for the batch size of 64, hence we lowered the batch size to 16 but kept the batch size at 64 for \acrshort{t5}-Small.
The task prefix we use is `classification ' for all the tasks, as the model takes a hyperparameter called a task prefix.

\section{Methods}
\label{methodsapp}

\subsection{Preprocessing}
\label{preprocess}

We carried out preprocessing on all the data to remove duplicates and unwanted strings or characters.
In some of the datasets, such as \acrshort{olid} (task C), there are "nans" (empty entries) in some columns of the labels.
These cause problems for the models by dropping model performance.
We, therefore, dropped such rows during the preprocessing step.
To prepare the text for the models, the following standard preprocessing steps are applied to all the datasets: 

\begin{itemize} 

\item URLs are removed. 
\item emails are removed.
\item IP addresses are removed.
\item Numbers are removed. 
\item All characters are changed to lowercase.
\item Excess spaces are removed. 
\item Special characters such as hashtags(\#) and mention symbols (\makeatletter @)  are removed. 

\end{itemize}

\subsection{Metrics}
\label{metrics}
The F1 score is the harmonic mean of the precision and recall.
The relative contribution to the F1 from precision and recall are equal.
We report both weighted and macro F1 scores because of past studies.
Macro-F1 does not take label imbalance into account unlike weighted-F1, which accounts for label imbalance by finding the labels' mean weighted by support (each label's true instances) \cite{scikit-learn}.\footnote{scikit-learn.org/../generated/sklearn.metrics.f1\_score.html}

\subsection{Augmentation}
\label{augment}

\subsubsection{Word-Level Deletion}
The first technique involves the use of the list of offensive words available from the online resource at Carnegie Mellon University.\footnote{cs.cmu.edu/\texttildelow biglou/resources/}
This list is used to ensure offensive tokens are not deleted during the pass through the training set.
From the original list of 1,383 English words, we removed 160 words, which we considered may not qualify as offensive words because they are nationalities/geographical locations or adjectives of emotions.
1,223 words are left in the document used for our experiment.
Examples of words removed are: \textit{european, african, canadian, american, arab, angry} and many unharmful words.
Samples ending or starting with offensive words are kept as they are in the new augmented training data and are therefore dropped when merged with the original, to avoid duplicates.

\subsubsection{Conversation Generation}
The second technique involves the use of the dialogue (conversation) model checkpoint by \citet{adewumi2021sm}, which was finetuned on the \acrfull{multiwoz} dataset by \citet{eric-EtAl:2020:LREC}.
It is an autoregressive model based on the pretrained \acrshort{dialogpt}-medium model by \citet{zhang2020dialogpt}.
An autoregressive model conditions each output word on previously generated outputs sequentially \cite{zou-etal-2021-thinking}.
Every sample from the training data is used as prompt to the model to generate a response, which is then concatenated to the prompt to form a new version of the prompt that was supplied.
This ensures the original label of each sample is unchanged, as the offensive content, if any, is still retained in the new sample.
As demonstrated by \citet{adewumi2022vector}, the possibility of generating an offensive token is small for this model because the \acrshort{multiwoz} dataset it is trained on is reputed to be non-toxic.

This second technique literally doubled the training set size of the \acrshort{hasoc} 2021 dataset.
Random quality inspection was carried out on the augmented data.
Examples of the original and augmented samples from the \acrshort{hasoc} 2021 dataset, using this second technique, are given in Table \ref{table:hasaug}.
The offensive words (masked with ***) are retained in the new samples.
The top-p and top-k variables of the decoding algorithm for the model were set as p=0.7 and k=100, respectively.
Additional hyperparameters include \textit{maximum decoding length}, set as 200 tokens, \textit{temperature}, set as 0.8, and \textit{maximum ngram repeat limit}, set as 3.
These hyperparameters are based on previous work, as they have been shown to perform well \cite{adewumi2021sm}.

\subsection{The Ensemble}
The ensemble is a majority-voting system comprising of \acrshort{t5}-Base, \acrshort{t5}-Small and \acrshort{roberta}-Base models.
The saved model checkpoint from each trained model is used to make prediction on each sample of the test set of the \acrshort{hasoc} 2021 dataset for subtask A.
The prediction ("HOF" or "NOT") with more than one vote (2 or 3) is recorded as the prediction for that sample.
The weighted and macro F1 scores are then calculated with the \textit{scikit-learn} library \cite{scikit-learn}, as in all other cases.

\subsection{Cross-Task Training}
We performed cross-task training to ascertain if there will be performance gains on a target subtask.
We discover cross-task training can improve performance, however, not always.
Six subtasks from three datasets are selected for this purpose because of time and resource constraints.
The subtasks are all subtasks A (binary classification) from 3 datasets: \acrshort{olid}, \acrshort{hasoc} 2020, and \acrshort{hasoc} 2021.
Only the \acrshort{t5} model is used for these experiments.
We finetune an initial (source) subtask and then further finetune on a final (target) subtask of a different dataset before evaluating the test set of the target subtask. 

\section{\acrshort{xai}}
\label{sec:xaialg}

\subsection{\acrfull{ig}}
\begin{figure*}[h!]
\centering
\includegraphics[width=0.9\textwidth]{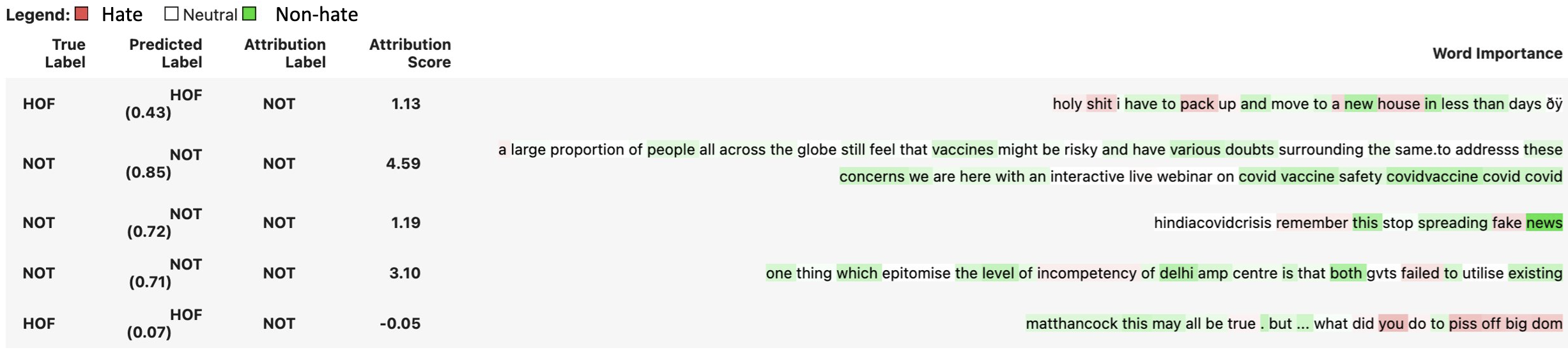}
\caption{Visualize attributions for \acrshort{bilstm} on \acrshort{hasoc} 2021 test set (correct-classification)}
\label{fig:xai1}
\end{figure*}

\begin{figure*}[h!]
\centering
\includegraphics[width=0.9\textwidth]{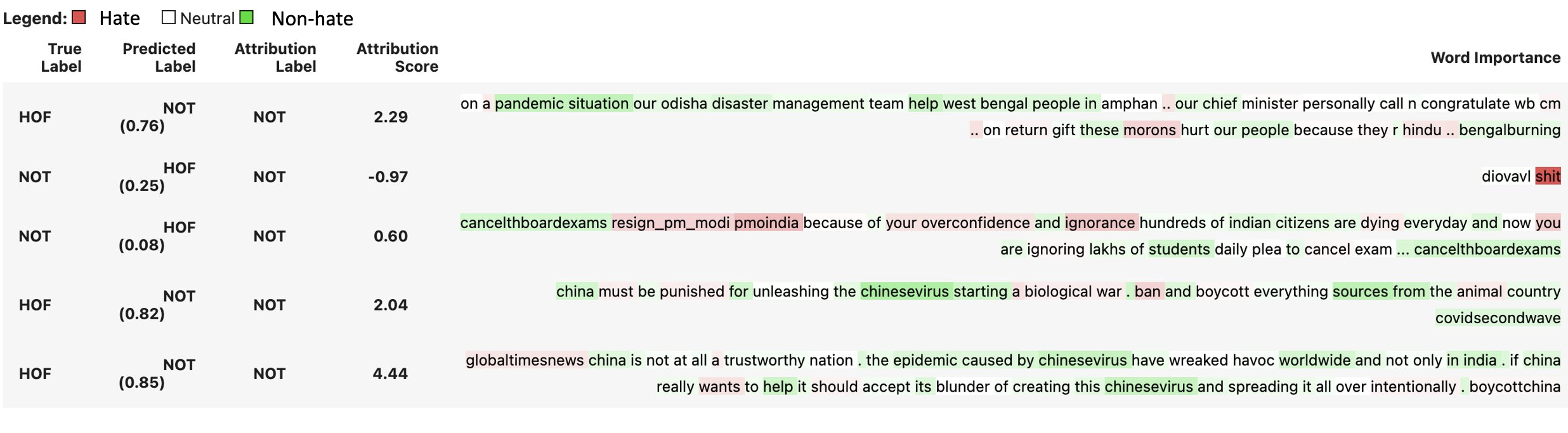}
\caption{Visualize attributions for \acrshort{bilstm} on Hasoc 2021 test set (miss-classification)}
\label{fig:xai2}
\end{figure*}

We apply \acrshort{ig} to the \acrshort{bilstm}.
It is an attribution method that is based on two fundamental axioms—
Sensitivity and Implementation Invariance \cite{sundararajan2017axiomatic}.
Generally, integrated gradients aggregate the gradients along the straight line between the baseline and the input.
A good baseline (of a zero input embedding vector, in this case) is very important.
Models trained using gradient descent are differentiable and \acrshort{ig} can be applied to these.
\acrshort{ig} has the advantage of being relatively faster than \acrshort{shap} computationally.
Section \ref{cherrylstm}, in the appendix, show \acrshort{ig} explanations for examples of 5 correctly classified (Figure \ref{fig:xai1}) and 5 incorrectly classified (Figure \ref{fig:xai2}) samples, based on the provided annotations.
The attribution shows which input words affect the model prediction and how strongly.
Important words are highlighted in shades of green or red, such that words in green contribute to non-hate speech while those in red contribute to hate speech. 
In Figure \ref{fig:xai2}, the second tweet has what may be considered an offensive word but it is incorrectly annotated as "NOT".
The \acrshort{bilstm}, however, predicts this correctly.

\subsection{\acrfull{shap}}
\acrshort{shap} assigns each feature an importance value for a particular prediction \cite{NIPS2017_7062}.
The exact computation of \acrshort{shap} values is challenging.
However, by combining insights from current
additive feature attribution methods, one can approximate them.
Its novel components include: (1)
the identification of a new class of additive feature importance measures, and (2)
theoretical results showing there is a unique solution in this class with a set of
desirable properties \cite{NIPS2017_7062}.
It unifies six existing methods: LIME, DeepLIFT, Layer-Wise Relevance Propagation, Shapley regression values, Shapley sampling values and Quantitative
Input Influence.
The last three are based on classic cooperative game theory \cite{shapley1951notes}.
This provides improved performance and consistency with human intuition.
\acrshort{shap} has the advantage that it can be applied to models whose training algorithm is differentiable as well as those based on non-differentiable algorithm, such as trees.

\begin{figure*}[h!]
\centering
\includegraphics[width=0.92\textwidth]{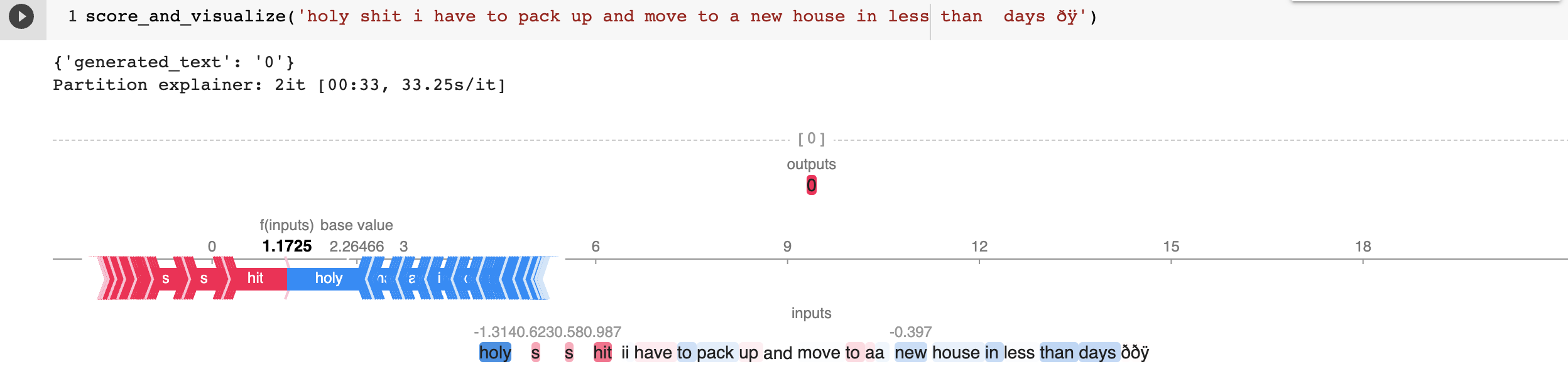}
\caption{\acrshort{shap} explanation of the \acrshort{t5} model prediction}
\label{fig:shap1}
\end{figure*}

\begin{figure*}[h!]
\centering
\includegraphics[width=0.92\textwidth]{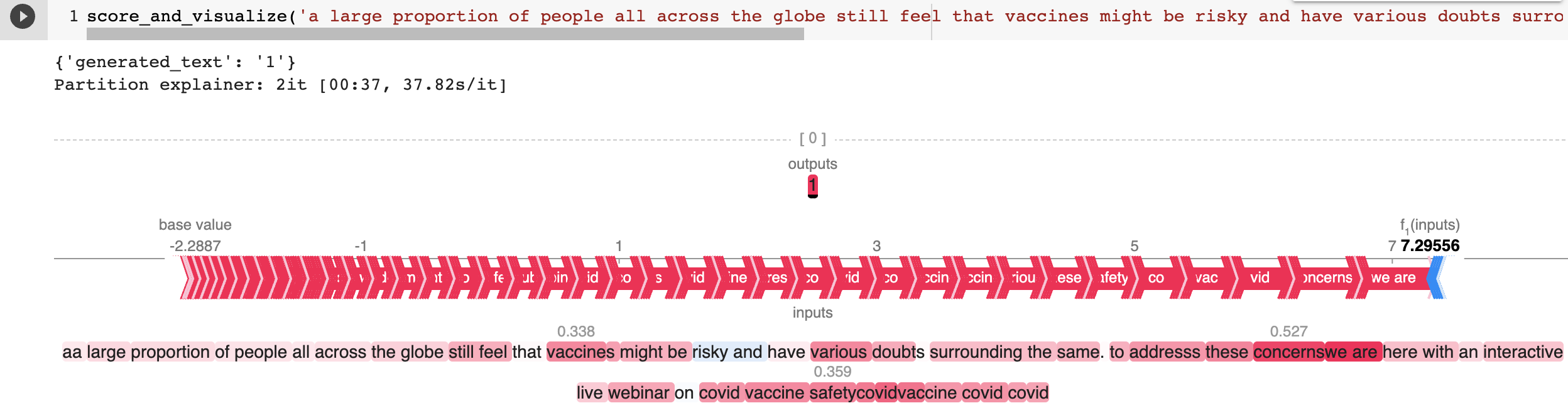}
\caption{\acrshort{shap} explanation of the \acrshort{t5} model prediction}
\label{fig:shap2}
\end{figure*}

\begin{figure*}[h!]
\centering
\includegraphics[width=0.92\textwidth]{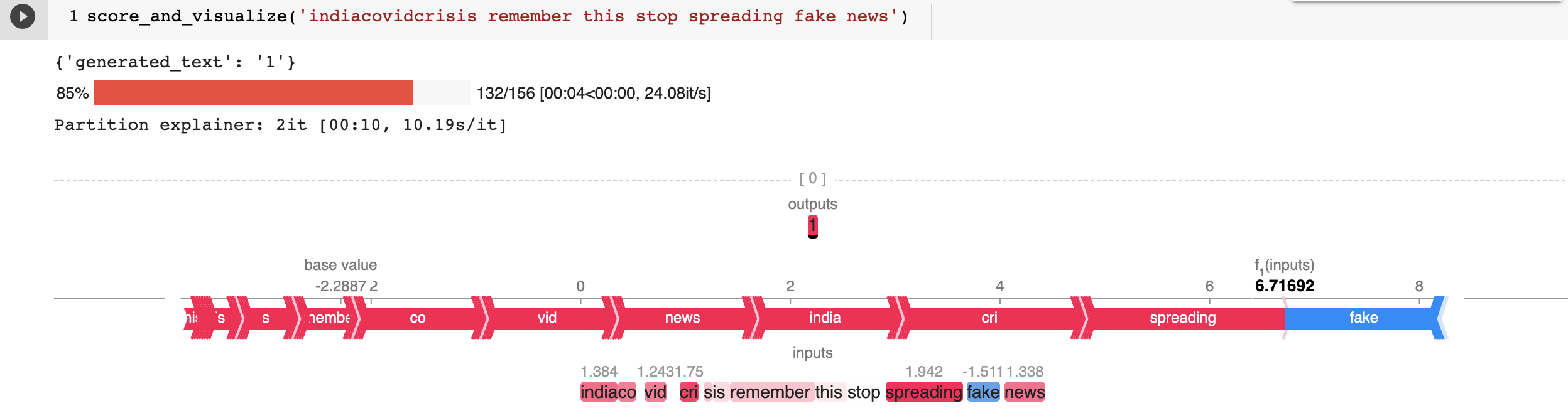}
\caption{\acrshort{shap} explanation of the \acrshort{t5} model prediction}
\label{fig:shap3}
\end{figure*}

\begin{figure*}[h!]
\centering
\includegraphics[width=0.92\textwidth]{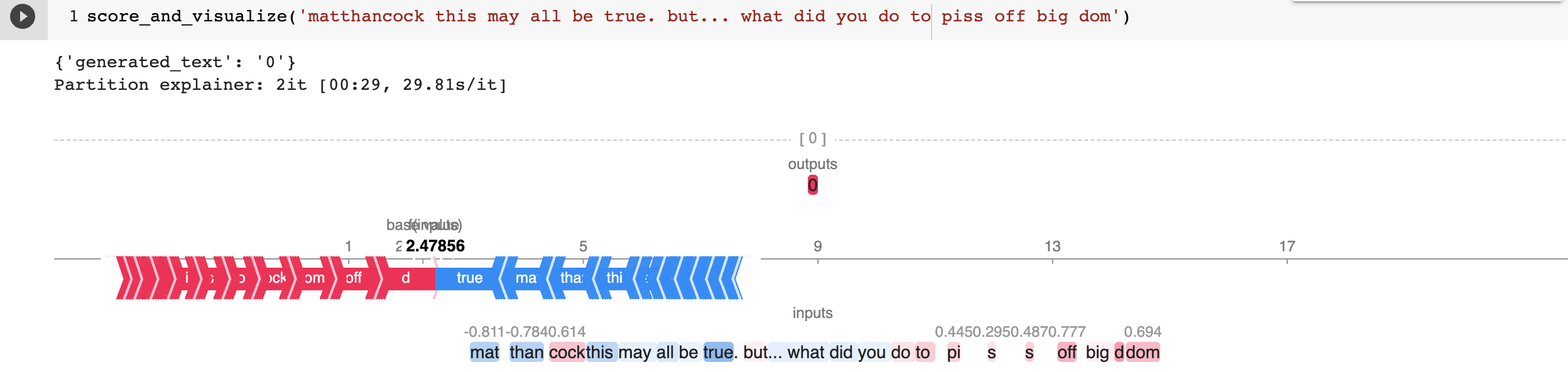}
\caption{\acrshort{shap} explanation of the \acrshort{t5} model prediction}
\label{fig:shap5}
\end{figure*}

\begin{figure*}[h!]
\centering
\includegraphics[width=0.92\textwidth]{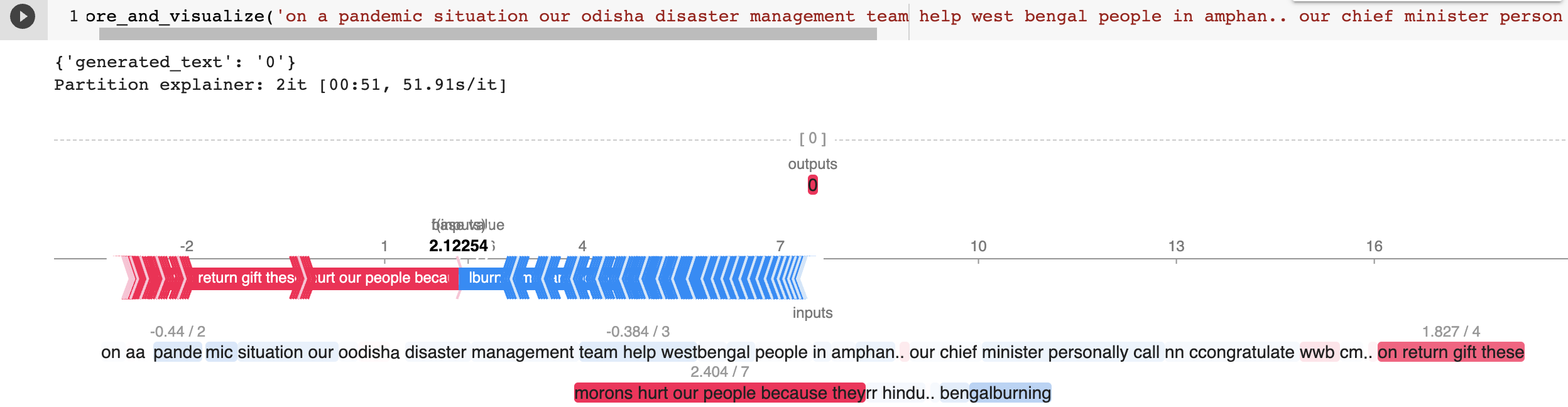}
\caption{\acrshort{shap} explanation of the \acrshort{t5} model prediction}
\label{fig:shap6}
\end{figure*}

\acrshort{shap} functionality is employed in this work by passing the supported HuggingFace transformers \acrshort{t5} pipeline (\textit{text2text-generation}) to \acrshort{shap}.
Important words or subwords are highlighted in shades of red or blue, such that words in red are those that contribute to a resulting prediction while those in blue contribute to what would be an alternative prediction.
The thicker the shade, the stronger the contribution, as also indicated by the real values above each word or subword.
Figures \ref{fig:shap1} to \ref{fig:shap6} present examples using the same samples from Table \ref{xai_samples}.
Additional examples are provided in Section \ref{cherryt5}, in the appendix.
We observe that 7 out of the 10 are correctly predicted by the \acrshort{t5}, as explained by \acrshort{shap}, compared to the 5 correct predictions by the \acrshort{bilstm}.

\subsection{\acrshort{shap} Explanations}

\section{Additional Results}
\label{sec:appendix}

\begin{table*}[h!]
\scriptsize
\centering
%\resizebox{\columnwidth}{!}{%
%\begin{tabular}{l|c|c}
\begin{tabular}{lcccc}
\hline
\textbf{Cross-task} &
\multicolumn{2}{c}{\textbf{Weighted F1 (\%)}} &
\multicolumn{2}{c}{\textbf{Macro F1 (\%)}}
\\
%\hline
 & Dev (sd) & Test (sd) & Dev (sd) & Test (sd)
\\
\hline
Hasoc 2020 A -> OLID A & 90.35 (0.01) & 83.94 (0.72) & 88.82 (0.01) & 79.81 (0.85)
\\
Hasoc 2021 A -> OLID A &  91.82 (0.01) & 83.52 (0.48) &  90.57 (0.01) & 79.22 (1.01)
\\
Hasoc 2021 A -> Hasoc 2020 A & 95.87 (0) & 90.14 (0.85) & 95.87 (0) & 90.13 (0.85)
\\
OLID A -> Hasoc 2020 A & 96.59 (0.68) & 91.73 (0.25) & 96.58 (0.68) & 91.73 (0.26)
\\ % \hline
OLID A -> Hasoc 2021 A  & 86.82 (0.01) & 80.91 (0.53) & 84.91 (0.02) & 79.32 (0.55)
\\ % \hline
Hasoc 2020 A -> Hasoc 2021 A & 87.2 (0.03) & 81.75 (0.29) & 87.37 (0.01) & 80.4 (0.3)
\\
\hline
\end{tabular}
%}
\caption{\label{table:crosst}Cross-Task inference using T5}
\end{table*}

\begin{table*}[h!]
\scriptsize
\centering
%\resizebox{\columnwidth}{!}{%
%\begin{tabular}{l|c|c}
\begin{tabular}{lcccc}
\hline
\textbf{Task} &
\multicolumn{2}{c}{\textbf{Weighted F1 (\%)}} &
\multicolumn{2}{c}{\textbf{Macro F1(\%)}}
\\
%\hline
 & Dev (sd) & Test (sd) & Dev (sd) & Test (sd)
\\
\hline
\textbf{Bi-LSTM} & \multicolumn{4}{c}{}
\\
\hline
HatEval SemEval 2019 A & - & 72.38 (0.54) & -  & 72.12 (0.72)
\\
HatEval SemEval 2019 B &  - & 77.74 (2.8) & - & 73.11 (0.44)
\\
Hasoc 2020 A &  88.6 (0.15) &  89.30 (0.15) & 89.47 (1.47)  & 90.28 (0.20)
\\ % \hline
Hasoc 2020 B & 75.80 (0.56) & 74.39 (2.31)  & 42.99 (0.15) & 42.97 (0.06)
\\
\acrshort{hso} & 90.19 (0.03) & - & 68.77 (1.93) & -
\\
Trolling, Aggression &  68.69 (0.36) & - & 36.00 (0.27)  & -
\\
\hline
\textbf{CNN} & \multicolumn{4}{c}{}
\\
\hline
HatEval SemEval 2019 A & - & 73.95 (0.64) & - & 71.67 (0.43)
\\
HatEval SemEval 2019 B & - & 78.88 (0.55) & - & 71.13 (0.43)
\\
Hasoc 2020 A &  88.06 (0.41) & 89.76 (0.44) & 88.21 (0.41)  & 90.08 (0.46)
\\ % \hline
Hasoc 2020 B &  76.38 (0.63 ) & 76.48 (0.61) &  49.15 (1.25) & 47.58 (0.85)
\\
\acrshort{hso} & 88.52 (0.62) & - & 71.27 (0.74) & -
\\
\acrshort{trac} &  71.01 (1.73) & - & 40.24 (0.43) & -
\\
\hline
\textbf{T5-Base} & \multicolumn{4}{c}{}
\\
\hline
HatEval SemEval 2019 A & - & 87.07 (4.81) & - & 86.52 (5.11)
\\
HatEval SemEval 2019 B & - & 99.93 (0) & - & 99.88 (0)
\\
%HOS & 92.2 (0.71) & - & 71.27 (0.74) & -
%\\
\acrshort{trac} & 80.84 (3.96) & - & 56.97 (8.34)  & -
\\
\hline
\textbf{HateBERT} \cite{caselli-etal-2021-hatebert} & \multicolumn{4}{c}{}
\\
\hline
HatEval SemEval 2019 A & - & - & - & 0.516 (0.007)
\\
\hline
\end{tabular}
%}
\caption{\label{tab:compare2} Model comparison of mean scores for other \acrshort{hs} datasets. \footnotesize{(sd: standard deviation; '-' implies no informaton available)}}
\end{table*}

\begin{table*}[h!]
\scriptsize
\centering
%\resizebox{\columnwidth}{!}{%
%\begin{tabular}{l|c|c}
\begin{tabular}{p{0.20\linewidth} p{0.55\linewidth}p{0.06\linewidth}p{0.06\linewidth}}%{lccc}
\hline
\_id & text & task\_1 & task\_2
\\
\hline
60c5d6bf5659ea5e55def8b3 & 
When you’re the main b*tch https:\//\//t.co\//HWlNpEUiwS
&
NOT & NONE
\\
\hline
60c5d6bf5659ea5e55df0242 & miya four creeps into every thought i have what the f*ck & NOT & NONE 
\\
\hline
60c5d6bf5659ea5e55defe58 & At least we’re being freed from the shambles of the evangelical, but d*mn y’all couldn’t just go w the flow? Y’all just had to find jebus through crystals and astrology? Smdh & NOT & NONE
\\ % \hline
\hline
60c5d6bf5659ea5e55defe24 & @kumarmbayar @Actor\_Siddharth He is mentally Upset, Why he's not talking about \#BengalBurning and why stupid talks about \@Tejasvi\_Surya & NOT & NONE
\\
\hline
60c5d6bf5659ea5e55df0210 & 
@amemoott @Ripple @UCT\_news @qz @coinbase @cz\_binance låve been holding XRP for years. On Coinbase and binance . If you not going to help shut the f*ck up
&
NOT & NONE
\\
\hline
60c5d6bf5659ea5e55def288 & 
Like this tweet for a DM SC: jayjayson4225 \#dmme \#horny \#hornydm \#nsfwtwitter \#nsfw \#twitterafterdark \#whitedick \#whitecock \#cockrate \#nudes \#naughtydm \#dmsopen \#bwc \#cock \#dick \#nsfwtwt \#porn \#sex \#dickpic \#dickpics \#cumslut \#cum \#slut \#whore \#hotwife
&
NOT & NONE
\\
\hline
60c5d6bf5659ea5e55defdc1 & 
happy p*ss off old people month!
&
NOT & NONE
\\
\hline
60c5d6bf5659ea5e55defc79 & 
\#China is such a worst country first they unleashed \#Chinesevirus on whole world and \#ChinaLiedPeopleDied and now india is struggling with \#Chinesevirus but they are mocking people who are dieing of \#ChinesVirus in India https:\//t.co\/V0AiuJV3lm
&
NOT & NONE
\\
\hline
%60c5d6bf5659ea5e55def1c7 & 
%The city with its masked inhabitants n deserted streets looks spooky!! \#ChineseVirus ka prakop·
%&
%NOT & NONE %- T5 also wrong
%\\
%\hline
60c5d6bf5659ea5e55df01aa & 
@globaltimesnews Communist must feel shame God is watching you what you did with world. \#CommunistVirus
&
NOT & NONE %- T5 also wrong
\\
\hline
60c5d6bf5659ea5e55def750 & 
@RapidWolfReturn @Utd\_76 @MenachoManuel1 @ITheKTrainI @UnitedStandMUFC Yeah... sh*t move, but as has been said; heat of the moment, stupid comment he probably doesn't really back.. should've just explained it, owned it, and moved on. He's a w*nker, regar
&
NOT & NONE
\\
\hline
\end{tabular}
%}
\caption{\label{table:hasissues}Hasoc 2021 test set samples with suspicious labels but correctly classified, in our view, as HOF by T5. \footnotesize
(parts of offensive words masked with "*")}
\end{table*}

\newpage

\begin{table*}[h!]
\scriptsize
\centering
%\resizebox{\columnwidth}{!}{%
%\begin{tabular}{l|c|c}
\begin{tabular}{p{0.20\linewidth} p{0.55\linewidth}p{0.06\linewidth}p{0.06\linewidth}}%{lccc}
\hline
\_id & text & task\_1 & task\_2
\\
\hline
60c5d6bf5659ea5e55def475 & 
holy shit i have to pack up and move to a new house in less than  days ðÿ
&
HOF & PRFN
\\
\hline
60c5d6bf5659ea5e55df026b & 
a large proportion of people all across the globe still feel that vaccines might be risky and have various doubts surrounding the same. to addresss these concerns we are here with an interactive live webinar on covid vaccine safety covidvaccine covid covid
&
NOT & NONE
\\
\hline
60c5d6bf5659ea5e55deff15 & 
indiacovidcrisis remember this stop spreading fake news
&
NOT & NONE
\\
\hline
60c5d6bf5659ea5e55defc3e & 
one thing which epitomise the level of incompetency of delhi amp centre is that both gvts failed to utilise existing healthcare infra built by earlier gvts to its capacity even failed to supply required oxygen this is sheer incompetency amp negligence. both should resign resignmodi
&
NOT & NONE
\\
\hline
60c5d6bf5659ea5e55df028c & 
matthancock this may all be true. but... what did you do to piss off big dom
&
HOF & PRFN
\\
\hline
60c5d6bf5659ea5e55defb7f & 
on a pandemic situation our odisha disaster management team help west bengal people in amphan.. our chief minister personally call n congratulate wb cm.. on return gift these morons hurt our people because they r hindu.. bengalburning
&
HOF & HATE
\\
\hline
60c5d6bf5659ea5e55defca7 & 
diovavl shit
&
NOT & NONE
\\
\hline
60c5d6bf5659ea5e55def240 & 
cancelthboardexams resign\_pm\_modi pmoindia because of your overconfidence and ignorance hundreds of indian citizens are dying everyday and now you are ignoring lakhs of students daily plea to cancel exam...cancelthboardexams
&
NOT & NONE
\\
\hline
60c5d6bf5659ea5e55defa7d & 
china must be punished for unleashing the chinesevirus starting a biological war. ban and boycott everything sources from the animal country covidsecondwave
&
HOF & HATE
\\
\hline
60c5d6bf5659ea5e55def5a2 & 
globaltimesnews china is not at all a trustworthy nation. the epidemic caused by chinesevirus have wreaked havoc worldwide and not only in india. if china really wants to help it should accept its blunder of creating this chinesevirus and spreading it all over intentionally. boycottchina
&
HOF & HATE
\\
\hline
\end{tabular}
%}
\caption{\label{xai_samples}Cherry-picked (pre-processed) examples from Hasoc 2021 test set for XAI in Sections \ref{cherryt5} \& \ref{cherrylstm}.}
\end{table*}

\subsection{Cherry-picked Examples from the \acrshort{hasoc} 2021 Test Set for \acrshort{t5} Explained by \acrshort{shap}}
\label{cherryt5}

\begin{figure*}[h!]
\centering
\includegraphics[width=0.92\textwidth]{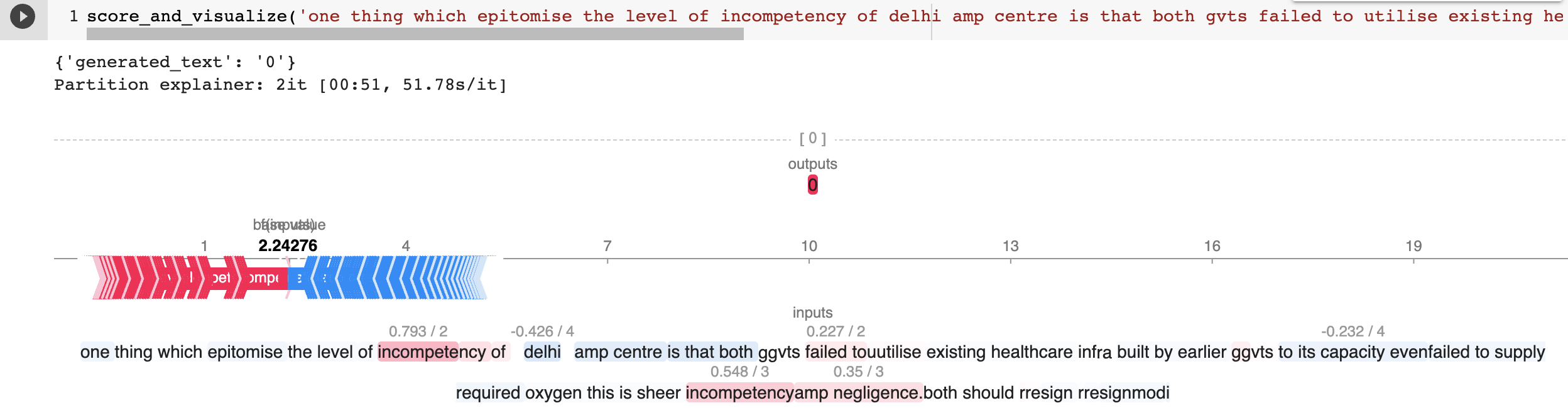}
\caption{\acrshort{shap} explanation of an incorrect \acrshort{t5} model prediction}
\label{fig:shap4}
\end{figure*}

\begin{figure*}[h!]
\centering
\includegraphics[width=0.92\textwidth]{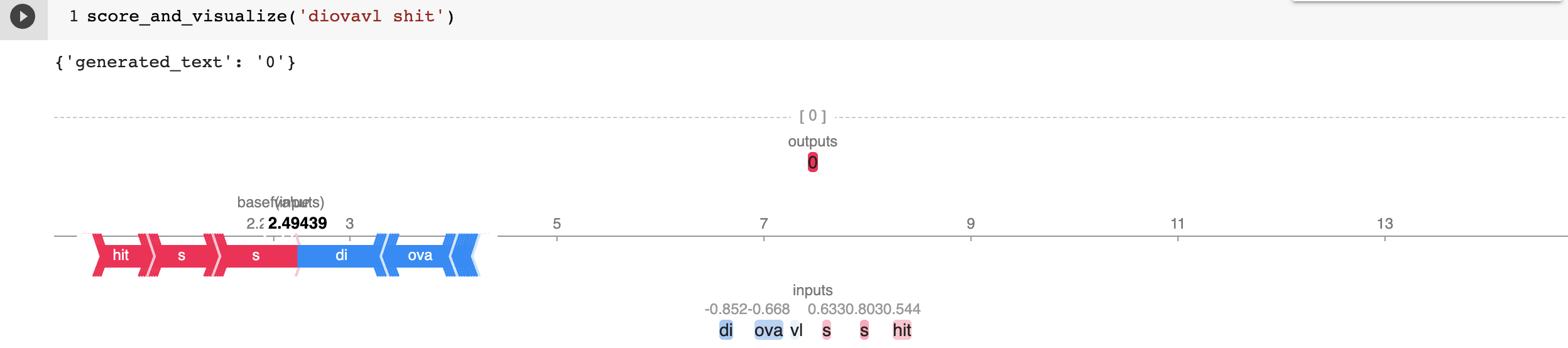}
\caption{\acrshort{shap} explanation of the \acrshort{t5} model prediction}
\label{fig:shap7}
\end{figure*}

\begin{figure*}[h!]
\centering
\includegraphics[width=0.92\textwidth]{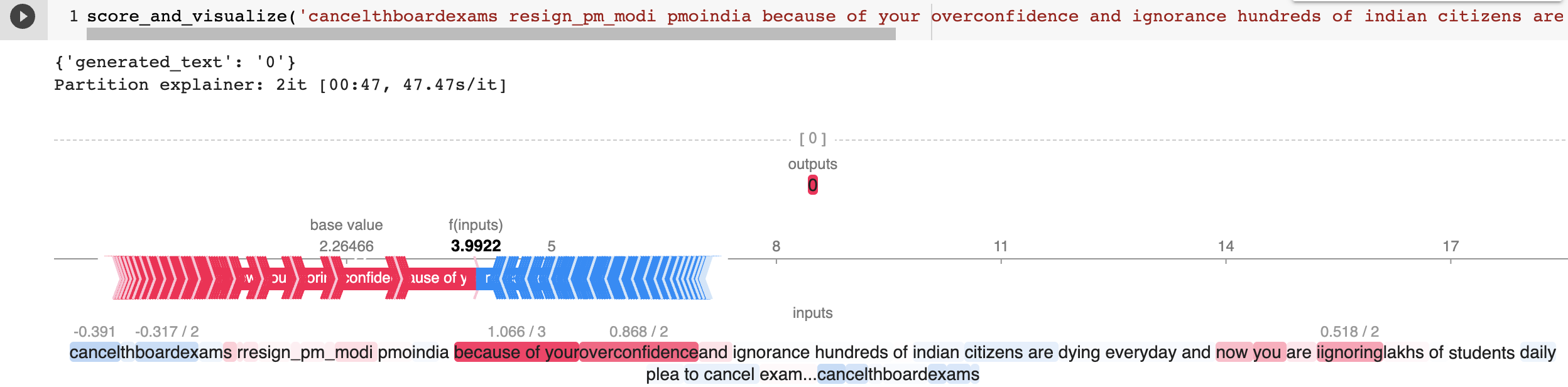}
\caption{\acrshort{shap} explanation of the \acrshort{t5} model prediction}
\label{fig:shap8}
\end{figure*}

\begin{figure*}[h!]
\centering
\includegraphics[width=0.92\textwidth]{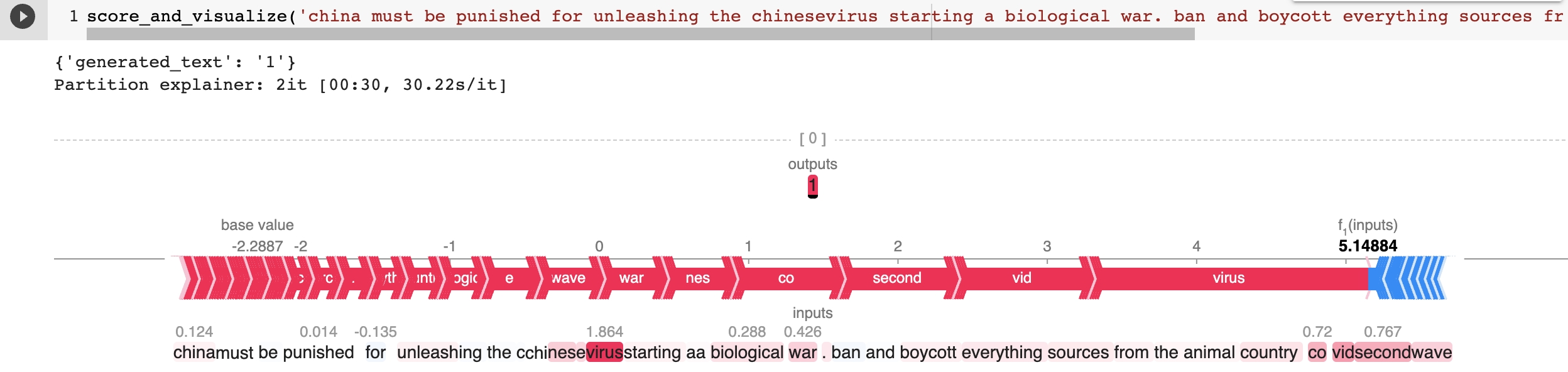}
\caption{\acrshort{shap} explanation of an incorrect \acrshort{t5} model prediction}
\label{fig:shap9}
\end{figure*}

\begin{figure*}[h!]
\centering
\includegraphics[width=0.92\textwidth]{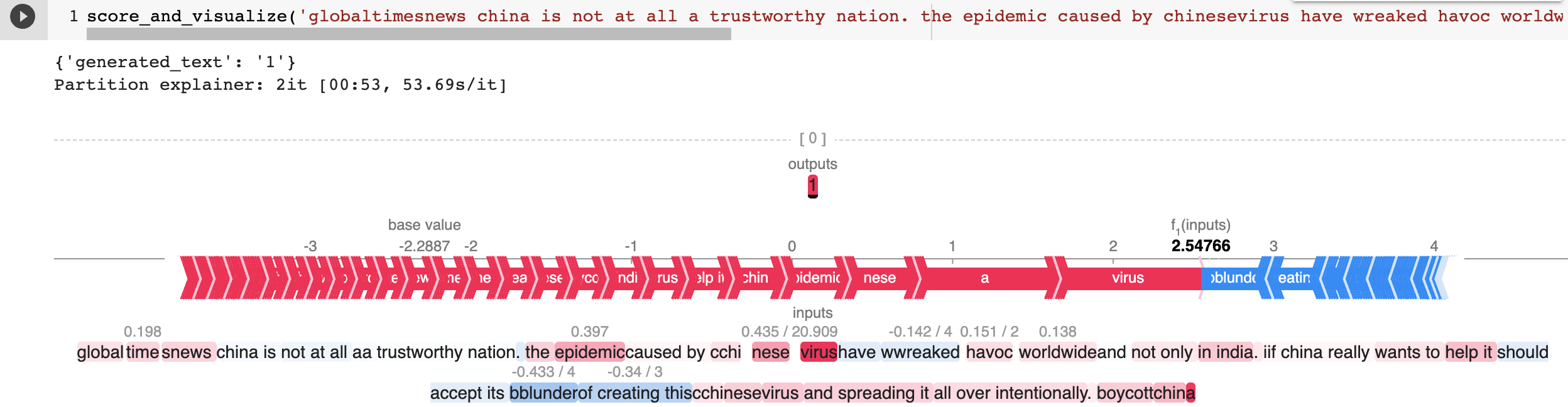}
\caption{\acrshort{shap} explanation of an incorrect \acrshort{t5} model prediction}
\label{fig:shap10}
\end{figure*}

\subsection{Cherry-picked Examples from the \acrshort{hasoc} 2021 Test Set for \acrshort{bilstm} Explained by \acrshort{ig}}
\label{cherrylstm}

%\begin{figure*}[h!]
%\centering
%\includegraphics[width=0.8\textwidth]{EMNLP 2022/wordratio.png}
%\caption{Word counts for "indiacovidcrisis" and "chinesevirus" hashtags in the \acrshort{hasoc} 2021
%training set}
%\label{fig:wordratio}
%\end{figure*}

\subsection{Some Incorrect \acrshort{hasoc} 2021 Annotations Correctly Classified by \acrshort{t5} \& Explained by \acrshort{shap}}
\label{incorrect}

\begin{figure*}[h]
\centering
\includegraphics[width=0.92\textwidth]{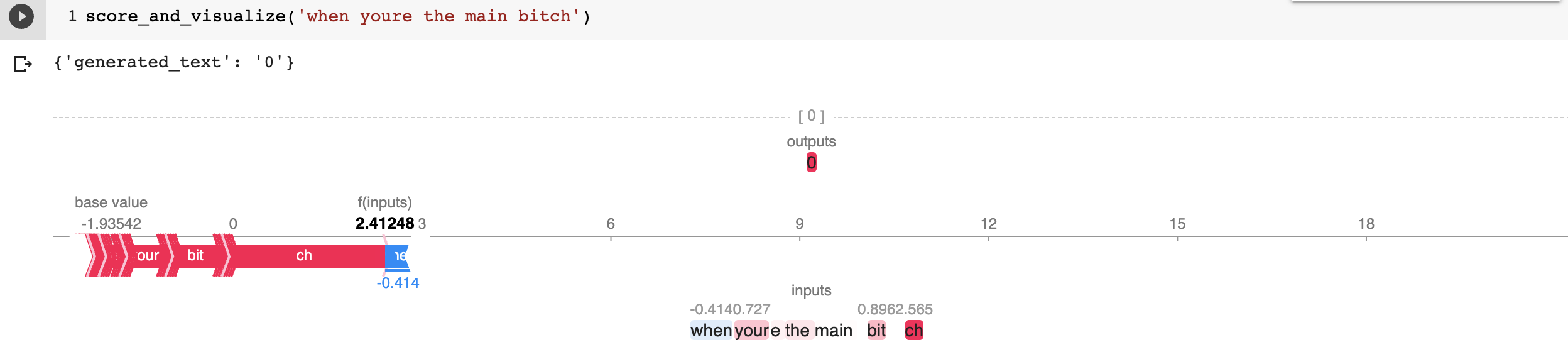}
\caption{Incorrectly annotated but correctly classified by \acrshort{t5}, as explained by \acrshort{shap}}
\label{fig:susp1}
\end{figure*}

\begin{figure*}[h]
\centering
\includegraphics[width=0.92\textwidth]{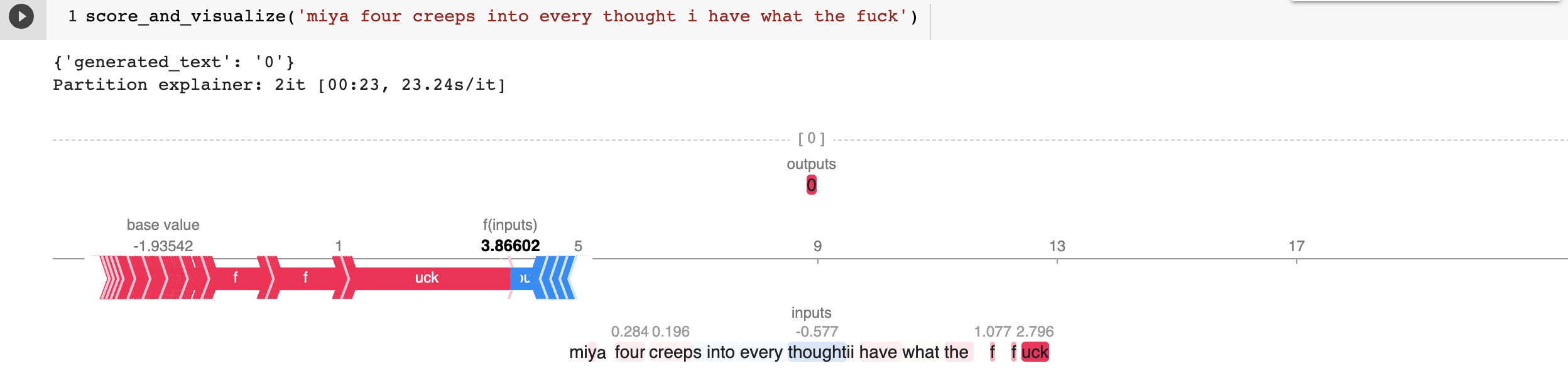}
\caption{Incorrectly annotated but correctly classified by \acrshort{t5}, as explained by \acrshort{shap}}
\label{fig:susp2}
\end{figure*}

\begin{figure*}[h]
\centering
\includegraphics[width=0.92\textwidth]{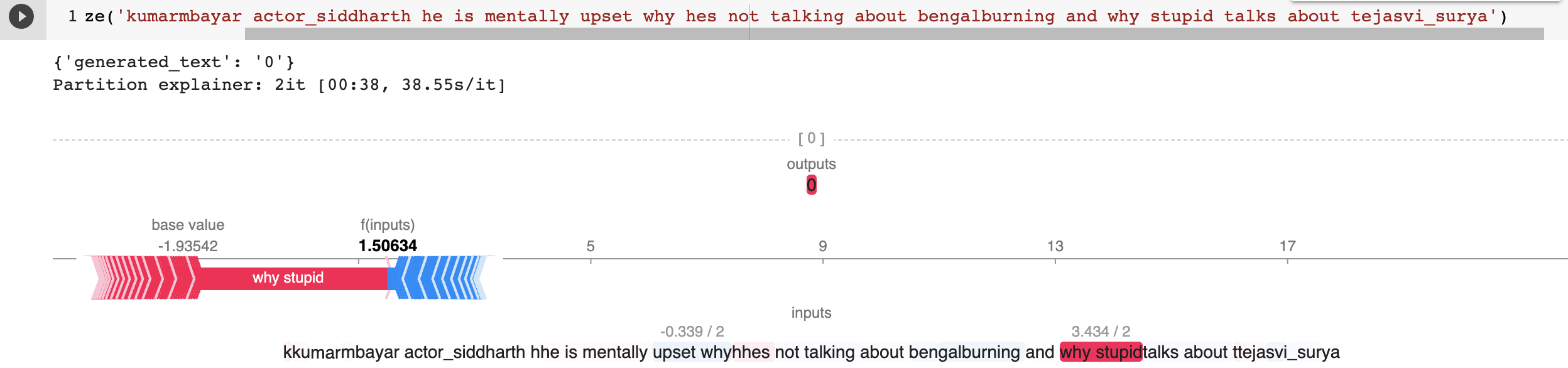}
\caption{Incorrectly annotated but correctly classified by \acrshort{t5}, as explained by \acrshort{shap}}
\label{fig:susp3}
\end{figure*}

\begin{figure*}[h]
\centering
\includegraphics[width=0.92\textwidth]{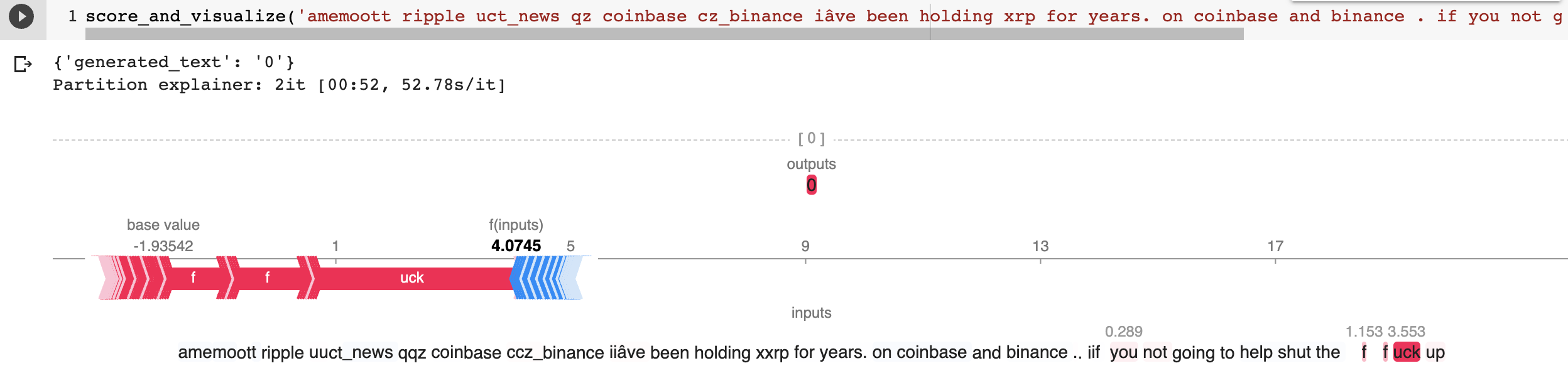}
\caption{Incorrectly annotated but correctly classified by \acrshort{t5}, as explained by \acrshort{shap}}
\label{fig:susp4}
\end{figure*}

\begin{figure*}[h]
\centering
\includegraphics[width=0.92\textwidth]{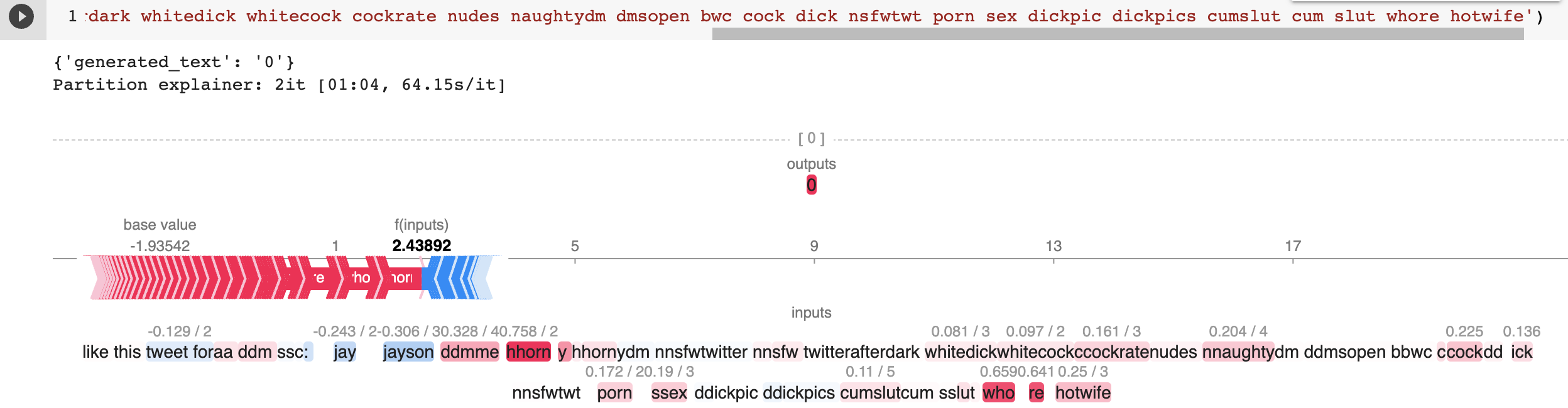}
\caption{Incorrectly annotated but correctly classified by \acrshort{t5}, as explained by \acrshort{shap}}
\label{fig:susp5}
\end{figure*}

\printglossary[type=\acronymtype]

\end{document}